\documentclass[twocolumn,letterpaper]{article}

\usepackage{xcolor}
\usepackage{graphicx}
\usepackage[accsupp]{axessibility} 
\usepackage{caption} 
\usepackage{lipsum} 

\newcommand{\fixed}[1]{\textcolor{black}{#1}}
\newcommand{\revised}[1]{\textcolor{black}{#1}}

\usepackage[pagenumbers]{cvpr} 

\definecolor{cvprblue}{rgb}{0.21,0.49,0.74}
\usepackage[pagebackref,breaklinks,colorlinks,citecolor=cvprblue]{hyperref}

\def\paperID{7734} 
\def\confName{CVPR}
\def\confYear{2024}

\title{LeGO: Leveraging a Surface Deformation Network for Animatable Stylized Face Generation with One Example}

\author{
Soyeon Yoon*\quad  Kwan Yun*\quad Kwanggyoon Seo\quad Sihun Cha\quad Jung Eun Yoo\quad Junyong Noh
 \vspace{3mm}
 \\KAIST, Visual Media Lab
 \vspace{3mm}
 \\{\href{https://kwanyun.github.io/lego/}{https://kwanyun.github.io/lego/}
}}

\begin{document}

\twocolumn[{
\maketitle
\begin{center}
    \begin{tabular}{c}
        \includegraphics[width=1\linewidth]{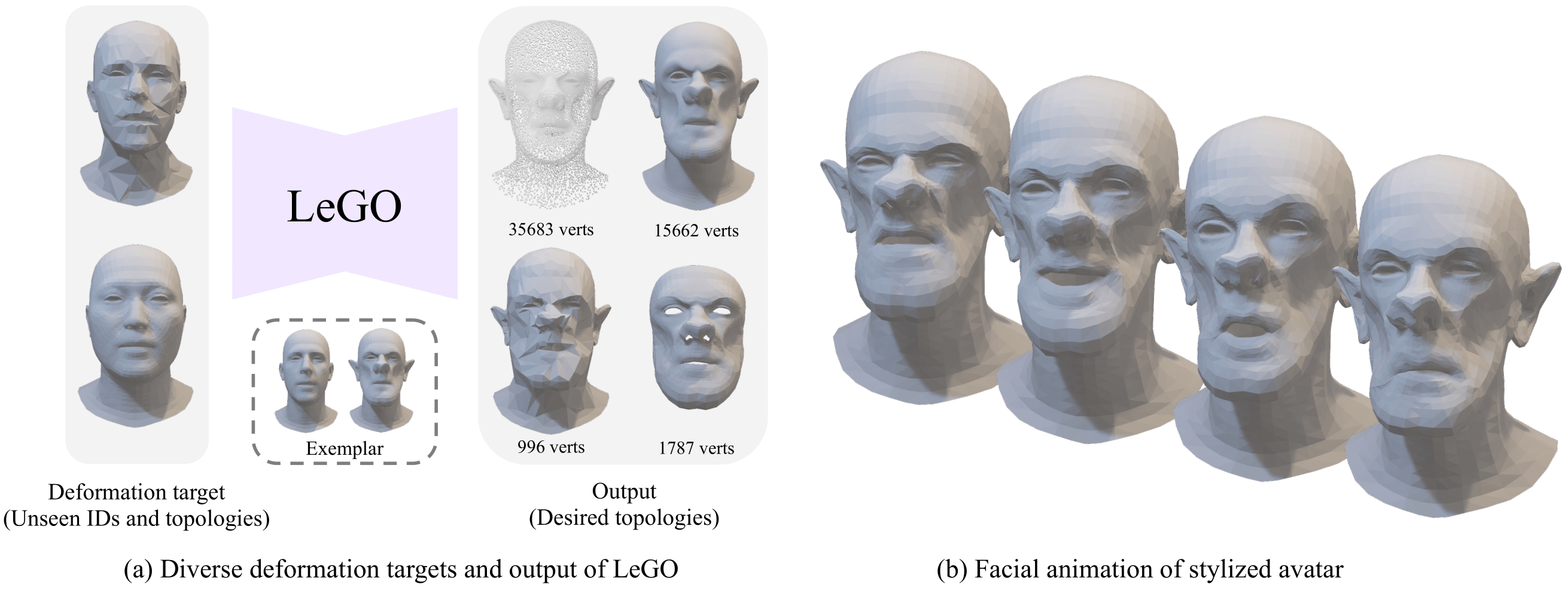}
    \end{tabular}
    \vspace{-2mm}
    \captionof{figure}{{(a) The proposed method demonstrates robustness to unseen face identities and topologies and effectively generates stylized output faces with desired topologies. (b) Our stylized avatars can be animated using 3DMM blend shapes.}}
    \label{teaser}
\end{center}
}]
\begin{abstract}
Recent advances in 3D face stylization have made significant strides in few to zero-shot settings. However, the degree of stylization achieved by existing methods is often not sufficient for practical applications because they are mostly based on statistical 3D Morphable Models (3DMM) with limited variations. To this end, we propose a method that can produce a highly stylized 3D face model with desired topology. Our methods train a surface deformation network with 3DMM and translate its domain to the target style using a paired exemplar.
The network achieves stylization of the 3D face mesh by mimicking the style of the target using a differentiable renderer and directional CLIP losses. Additionally, during the inference process, we utilize a Mesh Agnostic Encoder (MAGE) that takes deformation target, a mesh of diverse topologies as input to the stylization process and encodes its shape into our latent space.
The resulting stylized face model can be animated by commonly used 3DMM blend shapes.
A set of quantitative and qualitative evaluations demonstrate that our method can produce highly stylized face meshes according to a given style and output them in a desired topology. We also demonstrate example applications of our method including image-based stylized avatar generation, linear interpolation of geometric styles, and facial animation of stylized avatars. 
\end{abstract}
\vspace{-4mm}

\def\thefootnote{*}\footnotetext{These authors contributed equally to this work}\def\thefootnote{\arabic{footnote}}
\section{Introduction}
\label{sec:intro}

Crafting animatable stylized 3D avatars that encapsulate both personal identity and character style requires extensive efforts from skilled artists. When creating animated films, the artists design stylized 3D avatars whose facial appearance matches the theme of the entire content while putting careful effort into preserving the idiosyncrasy of the actors. Similarly, on social media, artists create numerous stylized presets so that the combinations of these presets can represent diverse identities. 

To reduce the burden of manual crafting effort, generating stylized 3D faces has been a prominent area of research. Recent attempts include 3D-aware generative adversarial networks (GANs) and denoising diffusion models (DMs)~\fixed{\cite{jin2022dr,abdal20233Davatargan,zhang2023styleavatar3d, kim2023Datid, li2023instruct}}, generating a stylized texture for the 3D morphable model (3DMM)~\cite{aneja2022clipface,zhang2023Dreamface}, text-based geometry deformation~\cite{gao2023textdeformer,michel2022text2mesh,mohammad2022clip, ma2023x,nguyen2023alteredavatar}, and surface deformation methods~\cite{yan2022cross,jung2022deep}. These methods have successfully demonstrated the possibility of producing high-quality and diverse stylized 3D faces, although each method has distinct challenges.

\begin{figure}[h]
\centering

\includegraphics[width=1\linewidth]{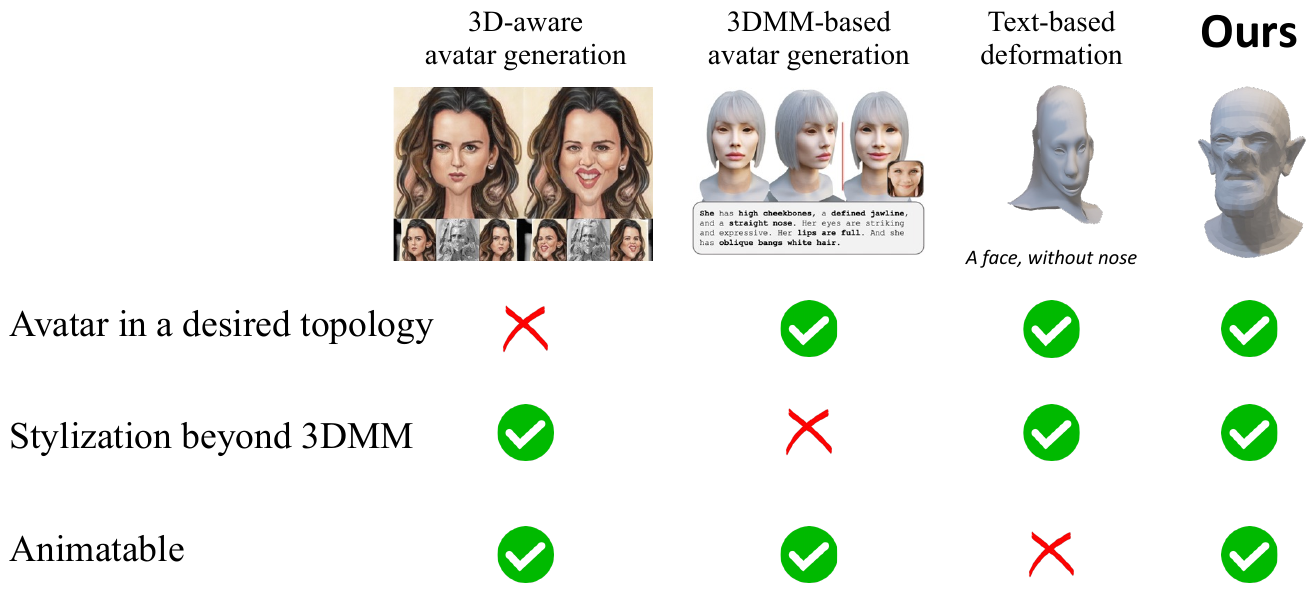}
\caption{Comparison of different stylized 3D face generation methods and their limitations in meeting key elements. 3D-aware methods cannot generate 3D face in desired topologies. 3DMM-based methods have a limited stylization capability. Text-based deformation models are not directly animatable. The proposed method meets the goal of all three components.}
\label{fig:intro}
\end{figure}

We identify three key elements for stylized avatar creation so that the results can be practically useful. 
\begin{enumerate}
\item \fixed{Avatar creation in a desired topology} that is compatible with conventional CG pipelines.
\item Extending stylization capabilities beyond 3DMM.
\item Generating stylized avatars that are animatable using blend shapes.
\end{enumerate}
Where each component meets the standards and demands of the dynamic entertainment landscape 1) allowing creators to reuse existing animation rigs and texture maps across different models.; 2) achieving greater diversity and flexibility in expressing unique and non-conventional characters.; 3) enabling animators to achieve coherent and natural movements across various facial features and expressions. While recent methods have made meaningful strides in one or two aspects of these key elements, there has not been a method that satisfies all three elements, as summarized in Figure~\ref{fig:intro}.
\vspace{0.5mm}
 
To address this, we propose a novel method that can generate stylized 3D face meshes. This is achieved by translating the domain of a pre-trained surface deformation network based on one of the most widely used 3DMM model, FLAME~\cite{li2017learning}, to a target style domain. We achieve this goal by first training the surface deformation network with the FLAME decoder to leverage its linear shape space combined with global expression space. During fine-tuning, we employ a directional CLIP-based domain adaptation method ~\cite{radford2021learning,zhu2022mind,kim2022dynagan}, widely used in 2D domain, to retain the face identity while reflecting the desired style.

In addition, to seamlessly integrate this 2D-based training method into the stylized 3D face generation task, we propose a hierarchical rendering scheme that captures local and global facial features, ensuring effective training and identity preservation.
In the inference stage, we introduce a Mesh Agnostic Encoder (MAGE) to enable mesh agnostic stylization for an input, which we call a deformation target that has various mesh topologies. MAGE \revised{is composed of} pre-trained encoders from Neural Face Rigging (NFR)~\cite{qin2023neural} and latent mapping networks, which establish correspondences between shapes by encoding mesh representations into a topology-invariant latent space.
This enhances the versatility and applicability of our approach in the context of 3D face stylization. As shown in Figure ~\ref{teaser}, our method can generate stylized 3D face models with varied mesh topologies that are equipped with the animation capability of 3DMM while ensuring consistency across diverse \revised{deformation target} mesh representations.

\section{Related Work}
\label{sec:formatting}

\subsection{Stylized 3D face generation}
With recent advances in the GANs and DMs that utilize neural fields for 3D-aware face generation \cite{gu2021stylenerf,chan2022efficient}, the generation of faces in diverse styles has gained popularity \fixed{\cite{abdal20233Davatargan,kim2023Datid,jin2022dr, li2023instruct}}. By leveraging 2D priors from generative models, \fixed{which capture various patterns and variations observed in the extensive 2D training data}, these methods can generate 3D faces with various styles. However, despite their success in producing consistent multi-view images through neural rendering, creating stylized faces in a desired topology is challenging, limiting their suitability for using existing graphics tools across different faces. \fixed{On the other hand}, 3DMM-based personalized 3D face generation methods with text-guidance~\cite{aneja2022clipface,zhang2023Dreamface} have shown success in producing high-fidelity stylized textures for 3D face models. However, these methods confine the shape of the generated face models within the 3DMM shape space, constraining possibilities for geometric exaggerations or abstraction beyond the training data.

\subsection{Learning-Based 3D Shape Network}
Recently, learning implicit functions for 3D shapes has demonstrated remarkable performance in representing complex geometric structures \cite{mescheder2019occupancy,niemeyer2019occupancy,peng2020convolutional,park2019deepsdf,gropp2020implicit}. In particular, DIF-Net \cite{deng2021deformed} adopted MLPs for learning a standard signed distance function (SDF) and a volumetric deformation function, leading to comprehensive mapping between the produced SDFs. Most related to our work, DD3C~\cite{jung2022deep} utilized template deformation \fixed{for 3D caricature auto-decoder. They found that modeling each shape as a deformation of a fixed temaplate surface effective compared to absolute position. We advance a step further by training a surface deformation network on 3DMM and transfer its domain into stylization.}

Another line of research aims to transfer shape deformation \cite{sumner2004deformation}. Recently, attempts have been made to transfer deformation utilizing examples \cite{yifan2020neural,yan2022cross,sung2020deformsyncnet} or text \cite{michel2022text2mesh,mohammad2022clip,gao2023textdeformer}. 
These methods can deform a mesh into a desired style and identity using a given example. However these deformation methods may not be as effective when it comes to coherent control of animation.

\subsection{Few-Shot Domain Adaptation}
Domain adaptation refers to the process of utilizing a neural network that has been initially trained on a large dataset from a source domain, and subsequently fine-tuning it on a smaller dataset from a target domain. Few-shot domain adaptation is widely researched in the 2D domain, especially with generative priors \cite{noguchi2019image,wang2020minegan,li2020few,ojha2021few,jin2022dr,chong2022jojogan,xiao2022few,yang2022pastiche}. Attempts have been made in one-shot domain adaptation to utilize the semantic capabilities of vision-language networks like CLIP \cite{radford2021learning}. These networks can provide direction for distinguishing between identity and style \cite{gal2022stylegan,zhu2022mind,kim2022dynagan}. We adopt this two-way directional guidance from CLIP and combine it with a differentiable renderer to effectively stylize the face mesh while preserving the original identity using \revised{a paired exemplar.}

\subsection{Mesh Agnostic Deformation}
Recently, mesh agnostic networks \cite{qi2017pointnet,qi2017pointnet++,groueix2018papier,hanocka2019meshcnn,sharp2022diffusionnet} demonstrated a great potential for learning 3D information such as dense correspondences between different 3D shape representations without relying on consistent topology or vertex ordering. These networks operate on meshes and encode shapes into a topology-invariant latent space.
Specifically, NFR~\cite{qin2023neural} proposed an autoencoder framework for facial expression retargeting across different mesh topologies. The framework utilizes separate expression and identity encoders, both functioning in a topology-agnostic manner.

We integrate components of this approach into our method by mapping their embeddings into our latent space. This network translates the output of the pre-trained encoder from NFR to the latent space of our surface deformation network. This enables any geometric shape of the \revised{deformation target} to be encoded and fed into our surface deformation network.

\section{Method}
\begin{figure*}[h]
\centering
\includegraphics[width=1\linewidth]{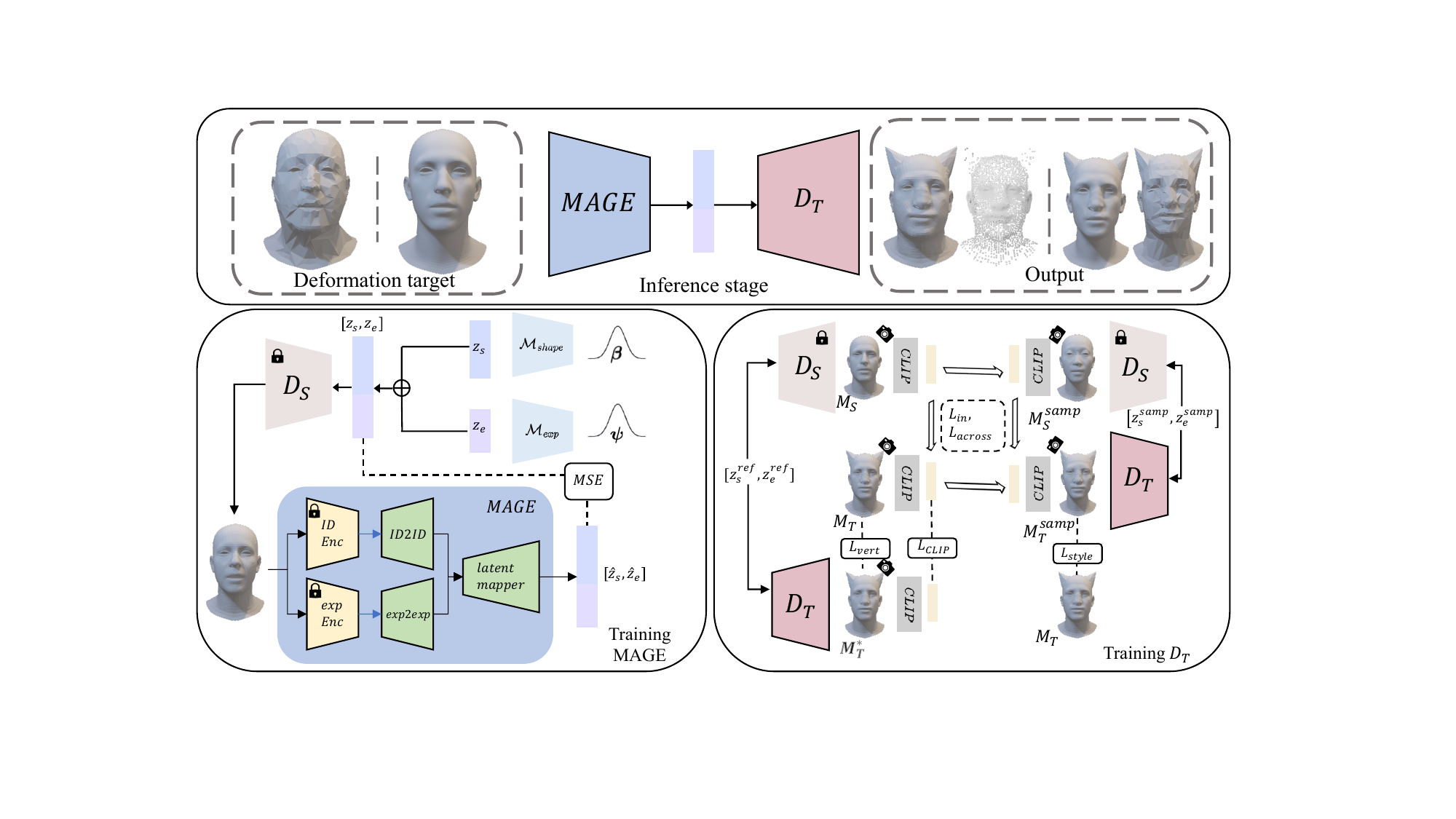}
\caption{Overview of our method: The upper box illustrates the inference stage, where our method takes diverse deformation targets and generates stylized outputs. In the lower-left box, the training process of Mesh Agnostic Encoder (MAGE) is depicted. In the lower-right box, the fine-tuning process of $D_T$ is illustrated.}
\vspace{-1em}
\label{method}
\end{figure*}

Our research focuses on training and fine-tuning a surface deformation network to generate stylized 3D faces with diverse shapes and expressions. 
We start with the source face deformation network $D_S$ that deforms the template face to a face with different identities and expressions. Thereafter, we fine-tune it into a target style face deformation network $D_T$ using \revised{a paired exemplar.} This process is outlined below.

\begin{enumerate}  
\item We first train $D_S$ using FLAME in a self-supervised manner, enabling the creation of versatile head meshes with different shapes and expressions.

\item For fine-tuning, we assemble \revised{a paired exemplar that consists of an identity exemplar mesh $M_S$ and a style exemplar mesh $M_T$, sharing identity.}

\item Both $M_{S}$ and $M_{T}$ jointly serve as example guidance to fine-tune $D_T$.

\item At inference, using MAGE and $D_T$, the \revised{deformation target} of diverse topologies can be translated into a stylized face.

\end{enumerate}
A detailed description of these processes will be provided in the following subsections.

\subsection{Deformation Network as Parametric Model}\label{subsec:pretrain}
$D_S$ is a surface deformation network that deforms a template face with the given latent vectors. We train the network to generate diverse human faces with FLAME, in a self-supervising manner. FLAME can manipulate both the global head shape and local expressions through its shape parameters  $\overrightarrow{\boldsymbol{\beta}} \in \mathbb{R}^{300}$, and expression parameters $\overrightarrow{\boldsymbol{\psi}} \in \mathbb{R}^{100}$, which are the components of the FLAME parameters $\Phi$. This empowers $D_S$ to generate diverse geometric face shapes and expressions.

To generate face meshes using these shape and expression components of $\Phi$ as input, we employ mapping networks, $\mathcal{M}_{shape}$ and $\mathcal{M}_{exp}$, which consist of Multi-Layer Perceptrons (MLP). Each mapping network separately transforms  $\overrightarrow{\boldsymbol{\beta}}$ and $\overrightarrow{\boldsymbol{\psi}}$ into latent vectors, $z_{s}$ and $z_{e}$, respectively. 
\begin{equation}
z_{s} = \mathcal{M}_{\text{shape}}(\overrightarrow{\boldsymbol{\beta}}), \quad z_{e} = \mathcal{M}_{\text{exp}}(\overrightarrow{\boldsymbol{\psi}})
\end{equation}
These latent vectors are then fed into $D_S$, enabling the creation of diverse face meshes imbued with a wide range of expressions. The mapping networks $\mathcal{M}_{\text{shape}}$ and $\mathcal{M}_{\text{exp}}$ are trained jointly with $D_S$ during the training process. The following loss function is employed:
\begin{equation}
L(\overrightarrow{\boldsymbol{\beta}}, \overrightarrow{\boldsymbol{\psi}}) = \left\| \text{FLAME}\left(\overrightarrow{\boldsymbol{\beta}}, \overrightarrow{\boldsymbol{\psi}}\right) - D_S\left(\begin{bmatrix} \mathcal{M}_{\text{shape}}(\overrightarrow{\boldsymbol{\beta}}) \\ \mathcal{M}_{\text{exp}}(\overrightarrow{\boldsymbol{\psi}}) \end{bmatrix}\right)\right\|^2_2
\end{equation}

\paragraph{Surface-Intensive Mesh Sampling}

To enable $D_S$ to represent \fixed{various geometries}, we introduce a surface-intensive mesh sampling (SIMS) strategy during training. SIMS is performed by randomly selecting points from the surfaces of the face mesh, \fixed{$
{FLAME}(\overrightarrow{\boldsymbol{\beta}}, \overrightarrow{\boldsymbol{\psi}})$,  during training $D_S$.} Specifically, we sample approximately 4 times more points from the surface than the number of vertices of the face mesh. Experiment results are reported in Sec.~\ref{exp:SIMS}.

\subsection{LeGO}\label{subsec:framework}
The fine-tuning process utilizes \revised{a paired exemplar}, $M_S$ and $M_T$, to guide the adaptation of $D_T$ for the generation of stylized 3D faces. \revised{An identity exemplar mesh}, $M_S$, is generated from the FLAME decoder using $\Phi_{ref}$, and \revised{a style exemplar mesh}, $M_T$, is manually crafted from $M_S$. During each iteration of fine-tuning,  $\Phi_{sample}$ is sampled randomly to generate face mesh $M_S^{samp}$ from $D_S$ and corresponding stylized 3D face $M_T^{samp}$ from $D_T$. In the fine-tuning process, quantities of $M_S$, $M_T$, $D_S$, and $D_T$ are all singular, representing one pair of meshes  and one pair of networks.

We introduce a hierarchical rendering scheme designed to preserve semantically important features, such as the shape and facial components, in the face mesh. With a differentiable renderer, this approach captures significant facial features from both local and global perspectives, enhancing stylization fidelity and identity preservation. To further enhance this process, we incorporate 2D-based losses, including CLIP reconstruction loss ${L}_{\mathrm{CLIP}}$, CLIP in-domain loss ${L}_{in}$, CLIP across-domain loss ${L}_{across}$, and 3D-based losses such as vertex reconstruction loss ${L}_{vert}$ and style loss ${L}_{style}$. Details of each loss are elaborated in Section~\ref{subsec:loss}.

\subsection{Loss Functions}\label{subsec:loss}

\paragraph{Vertex Reconstruction Loss} The vertex reconstruction loss ${L}_{vert}$ guides $D_T$ in learning to deform a \revised{style exemplar mesh} ${M_T}$ from \fixed{$[z_s^{ref},z_e^{ref}]$}, which is mapped from $\Phi_{ref}$. The loss utilizes Mean Squared Error (MSE) to ensure that the vertices of the predicted mesh ${M_T}^*$, generated by $D_T$, closely match $M_T$. The loss can be written as follows:
\begin{equation}
\label{eq:recon}
{L}_{vert} = \| M_{T} -{{M}_T}^* \|_2^2
\end{equation}
 
\vspace{-4.5mm} 
\paragraph{CLIP Reconstruction Loss} The CLIP reconstruction loss ${L}_{\mathrm{CLIP}}$ serves to maintain semantic consistency between the deformed mesh ${{M}_T}^*$ and $M_{T}$ in the CLIP space. This is crucial because even minor displacements in 3D space can result in surface irregularities or undesirable shading variations. The CLIP reconstruction loss can be written as follows:
\vspace{-5mm} 

\begin{equation}
\small
{L}_{\mathrm{CLIP}}=\sum_{l \in L} \sum_{v \in V_l}\left\|E_{C}\left(R_{l, v}\left(M_T\right)\right)-E_{C}\left(R_{l, v}\left({{M}_T}^*\right)\right)\right\|_2^2
\end{equation}

$R_{l,v}$ represents the hierarchical rendering of a differentiable renderer from level $l$ and view direction $v$. Each view is anchored at a predefined position on a face mesh, including the front and both sides of the face. Each level corresponds to the distance from the face to the camera. At the highest level, where the face is captured in close-up, additional images are rendered from significant facial features such as the nose, eyes, and lips. $E_{C}$ encodes rendered images into CLIP embeddings.

\paragraph{CLIP Directional Loss} Inspired from one-shot stylization methods in the 2D domain \cite{zhu2022mind}, we incorporate the CLIP in-domain loss ${L}_{in}$ and the CLIP across-domain loss ${L}_{across}$. ${L}_{in}$ ensures that the direction of two distinct faces in the source domain remains consistent in two distinct stylized faces of the target style domain. In contrast, ${L}_{across}$ ensures that the direction of a face from the source domain and its corresponding stylized face in the target domain is preserved across different faces and their corresponding stylized faces. This can be shown in the lower right of Figure~\ref{method}. Therefore our losses for CLIP in-domain and across-domain can be written as follows:
\vspace{1mm} 
\begin{equation}
\label{eq:clip_in}
\begin{aligned}
{L}_{\text{in}} &= \sum_{l \in L} \sum_{v \in V_l} \left\| (E_{C}(R_{l,v}(M_S^{samp}))- E_{C}(R_{l,v}(M_S))) \right. \\
&\quad - \left. (E_{C}(R_{l,v}(M_T^{samp})) - E_{C}(R_{l,v}(M_{T}))) \right\|_2^2
\end{aligned}
\end{equation}
\begin{equation}
\label{eq:across}
\begin{aligned}
{L}_{\text{across}} &= \sum_{l \in L} \sum_{v \in V_l} \left\| (E_{C}(R_{l,v}(M_T)) - (E_{C}(R_{l,v}(M_S))) \right. \\
&\quad - \left. (E_{C}(R_{l,v}(M_T^{samp})) - E_{C}(R_{l,v}(M_S^{samp}))) \right\|_2^2
\end{aligned}
\end{equation}
where $M_S^{samp}$ and $M_T^{samp}$ refer to $D_S[z_s^{samp};z_e^{samp}]$ and $D_T[z_s^{samp};z_e^{samp}]$, respectively, where $[z_s^{samp};z_e^{samp}]$ is the latent vectors mapped from $\Phi_{sample}$.

\vspace{-2mm}
\paragraph{Style Loss} \fixed{We introduce a novel style loss ${L}_{style}$ to capture the style of $M_T$ by utilizing surface normals. Style loss compares surface normal, which are strongly correlated with the semantic information of the mesh~\cite{deng2021deformed}. A straightforward approach would involve comparing $M_T$ with ${M_T}^{samp}$. However, this will constrain ${M_T}^{samp}$ to have same expression as $M_T$, which could degrade the animatability. To address this, we construct a pseudo pair for normal calculations. This is done by generating a face mesh through the concatenation of $z_e^{ref}$ from $M_T$ and $z_s^{samp}$. This approach is utilized to compute the style loss, ${L}_{style}$ without degrading the animatability while enhancing style adherence. In short, ${L}_{style}$ ensures alignment of the surface normals from $D_T([z_s^{samp};z_e^{ref}])$ with $M_T$.} The formulation of the style loss can be written as follows:

\begin{equation}
\label{eq:surface_normal}
{\small
{L}_{\text{style}} =\sum_{f \in S}{\left(1 - \frac{n_{f} \cdot {n'}_{f}}{|n_{f}| | {n'}_{f}|}\right)}
}
\end{equation}where $n_{f}$ refers to the surface normal from $M_T$ while ${n'}_{f}$ refers to the corresponding normal from $D_T([z_s^{samp};z_e^{ref}])$. 

The final objective function can be written as follows:
\begin{equation}
\label{eq:total}
\begin{split}
{L}_{total} = & \lambda_{vert} {L}_{vert} + \lambda_{\mathrm{CLIP}} {L}_{\mathrm{CLIP}} + \lambda_{in} {L}_{in} \\
& + \lambda_{across} {L}_{across} + \lambda_{style} {L}_{style}
\end{split}
\end{equation}
where $\lambda_{vert}$, $\lambda_{\mathrm{CLIP}}$, $\lambda_{in}$, $\lambda_{across}$, and $\lambda_{style}$ are the weight for each loss term.

\subsection{Mesh Agnostic Encoder}\label{subsec:pretrain}

NFR~\cite{qin2023neural} employs DiffusionNet \cite{sharp2022diffusionnet} for encoding and decoding face meshes, facilitating face retargeting across different topologies. Given NFR's proficiency in extracting identity and expression details from faces with varying topologies, we extend its encoders to create a MAGE, as depicted in Figure~\ref{method}. MAGE consists of ID2ID and exp2exp, both being MLPs that receive embeddings from each ID encoder and expression encoder pre-trained in NFR. Finally, \fixed{given intermediate vectors from both ID2ID and exp2exp}, the latent mapper outputs the latent vectors for $D_S$.

MAGE is trained in a self-supervised manner by randomly sampling $\beta$ and $\psi$ and passing them through the mapping networks of $D_S$ to obtain \fixed{ $[z_s;z_e]$. Training involves comparing $[\hat{z_s};\hat{z_e}]$, predicted from MAGE, with their corresponding ground truth latent vectors $[z_s;z_e]$.} The objective function of the encoder uses a MSE loss as follows:

\begin{equation}
\fixed{
{L}_{\text{enc}} = \left\| [z_{s};z_{e}] - [\hat{z_s};\hat{z_e}]\right\|_2^2
}
\label{eq:enc}
\end{equation}
where \revised{$[\hat{z_s};\hat{z_e}]$ is $\text{MAGE}\left[D_S{\left([z_{s};z_{e}]\right)} \right]$.} With this encoder, we can project a face mesh with diverse topologies into the latent space of $D_S$.

\subsection{Inference}\label{subsec:infer}

\revised{After training, $D_T$ is capable of generating a stylized 3D face mesh with a desired topology from a deformation target mesh that has an arbitrary topology.} This is accomplished by first projecting \revised{the deformation target} into a \fixed{latent vector}, using MAGE. \fixed{This latent vector} is then fed as input to $D_T$, allowing it to deform the template face of the desired topology into a stylized 3D face mesh. The resulting stylized face preserves both the identity of the \revised{deformation target} and the desired style while providing animation control.

\section{Experiments}
\label{sec:experiments}

\subsection{Qualitative Evaluation}
We evaluate our method based on three key elements for stylized 3D head avatar creation: Avatar creation in a desired topology, stylization beyond 3DMM, and animation capability with blend shapes. Also, implementation details are provided in supplementary material.

\begin{figure}[h]
\centering
\includegraphics[width=1\linewidth]{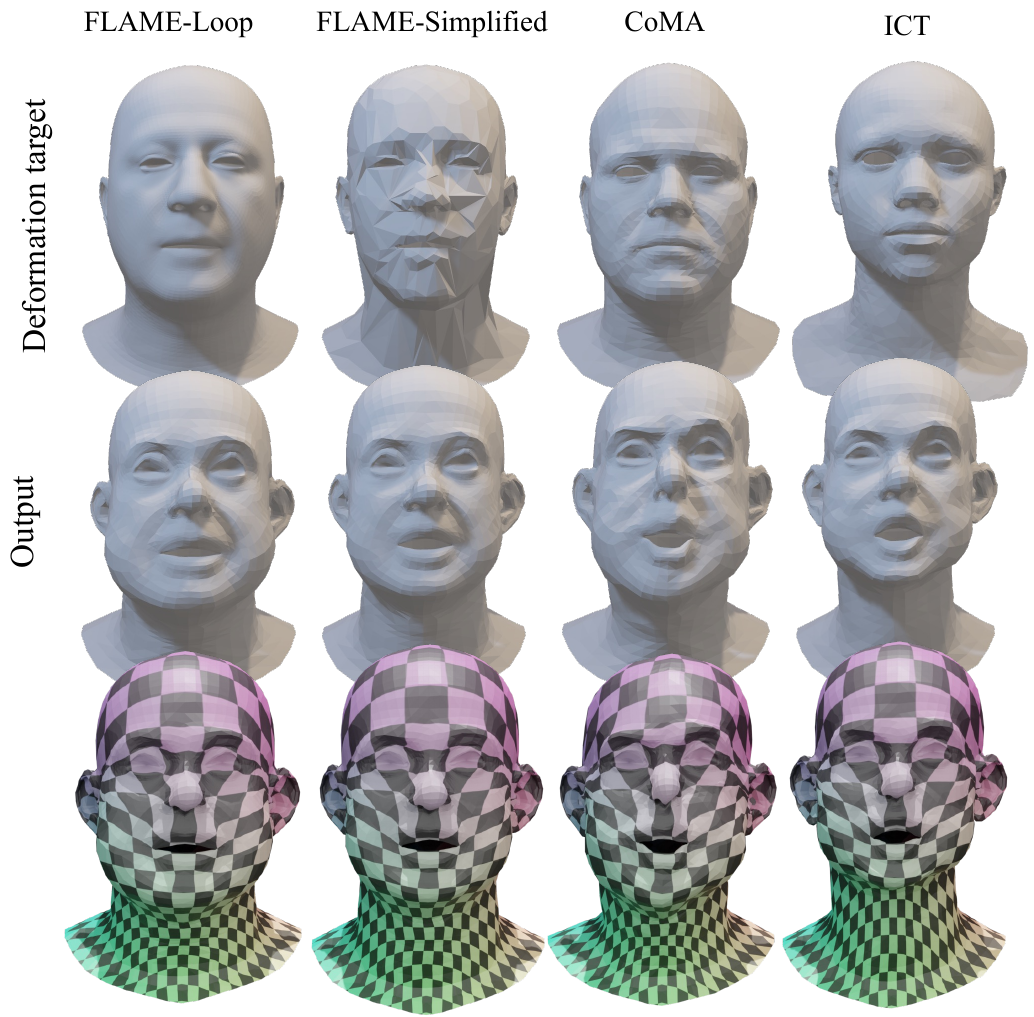}
\caption{Demonstration of stylized 3D faces with desired topology, regardless of deformation target variations.}
\vspace{-5pt}
\label{key1}
\end{figure}

\vspace{-2mm}
\paragraph{Avatar in a Desired Topology} 
\revised{To evaluate the capability of our method in producing a mesh with a desired topology from the deformation target of arbitrary topologies, we first apply a re-meshing technique that includes Loop subdivision~\cite{loop1987smooth} and mesh simplification~\cite{schroeder1992decimation} to meshes from FLAME.} 
We also use original meshes from the CoMA~\cite{ranjan2018generating} and ICT-FaceKit~\cite{li2020learning} directly as additional diverse meshes.
To visually confirm that our method can align different mesh topologies with a desired template face, we enhance the resulting meshes with a checkerboard texture. The visual results, as shown in Figure~\ref{key1}, demonstrate that our method consistently produces stylized 3D faces with the desired topology, regardless of variations in the topology of the \revised{deformation target}.

\begin{figure}[h]
\centering
\includegraphics[width=1\linewidth]{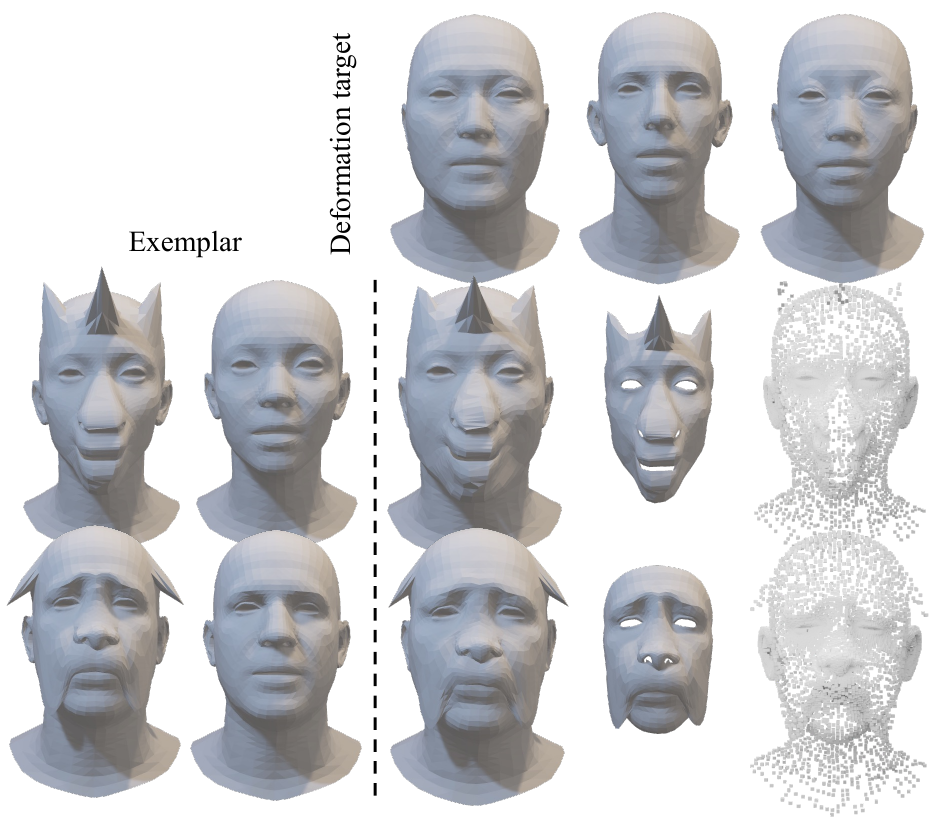}
\vspace{-20pt}
\caption{Stylization results across diverse styles and identities. Our approach generates varied styles while preserving the deformation target identity and generalizing to diverse geometric representations like masks and point clouds.}
\vspace{-5pt}
\label{key2}
\end{figure}

\vspace{-5mm} 
\paragraph{Stylization Capability}
We \fixed{evaluate our method across various styles to assess its stylization capability.} As shown in Figure~\ref{key2}, our method can generate a broad spectrum of styles encompassing human-like and non-human-like faces while preserving the original identities of the given face mesh. \fixed{Our method also demonstrates the capability to achieve stylization across various geometries, including face masks and point clouds.}

\begin{figure}[h]
\centering
\includegraphics[width=1\linewidth]{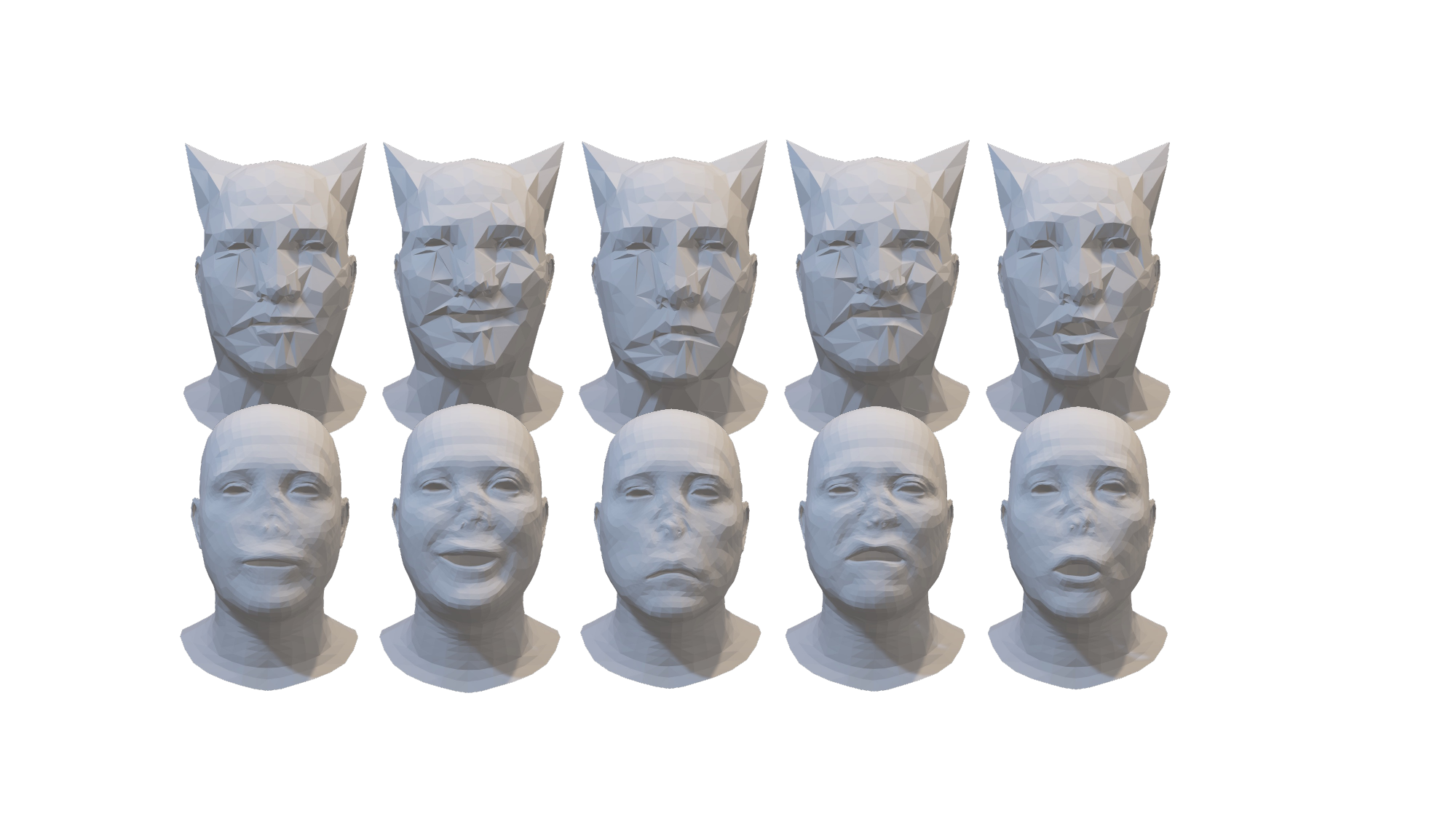}
\vspace{-12pt}
\caption{Visualization of dynamic expressions in stylized 3D faces.}
\vspace{-12pt}
\label{key3}
\end{figure}

\paragraph{Animation Capability}
Using parameters from 3DMM, our method can \fixed{animate facial expressions in the resulting stylized 3D faces.} As illustrated in Figure~\ref{key3}, our method can generate 3D stylized faces with \fixed{various expressions}. This allows applications such as video-driven stylized talking heads, which are elaborated in supplementary material.

\subsection{Quantitative Evaluation}

For a quantitative evaluation, we measured the average CLIP style preservation (CLIP-SP) and CLIP identity preservation (CLIP-IP) scores. These metrics reflect the trade-off between preserving style and identity, making their averages crucial for validating stylization. For both metrics, we calculated the cosine similarity of CLIP embeddings from rendered meshes. Specifically, for CLIP-SP, we compared the embeddings of the generated stylized face mesh and \revised{style exemplar mesh}. For CLIP-IP, we compared the embeddings of the generated stylized face mesh and \revised{deformation target mesh}. For dataset, we sampled 8 different face mesh from FLAME without expression and manually crafted stylized mesh corresponding to each identities. Also, for the quantitative evaluation, we randomly sampled another 10 different identities without expression to be used as \revised{deformation targets}.

\begin{table*}
\centering
\renewcommand{\arraystretch}{0.9} 
\setlength{\tabcolsep}{3pt}
\caption{Quantitative results are presented for comparison with baselines and ablations, featuring averages of CLIP-SP and CLIP-IP. The highest scores are denoted in bold, while the second highest scores are underlined.} \vspace{-10pt}
\small
\begin{tabular}{|c|ccc|ccc|ccc|c|}
\noalign{\smallskip}\noalign{\smallskip}\hline
Mesh Type & \multicolumn{3}{c|}{FLAME} & \multicolumn{3}{c|}{Loop} & \multicolumn{3}{c|}{Simplified} &Overall-Average\\ \hline
Methods & CLIP-SP & CLIP-IP &Average & CLIP-SP & CLIP-IP &Average& CLIP-SP & CLIP-IP &Average&Average\\
\hline
Ours & 0.9867 & 0.9347 & \textbf{0.9607} & 0.9833 & 0.9350 & \underline{0.9592} & 0.9559 & 0.8987 & \underline{0.9273} & \textbf{0.9491}\\ \hline
Ours w/o $L_{style}$ & 0.9809 & 0.9374 & 0.9591 & 0.9768 & 0.9339 & 0.9553 & 0.9542 & 0.8987 & 0.9265 & 0.9470 \\
Ours w direct $L_{style}$ & 0.9865 & 0.9348 & \textbf{0.9607} & 0.9833 & 0.9356 & \textbf{0.9595} & 0.9550 & 0.8973 & 0.9262 & \underline{0.9488} \\
w/o hierarchical rendering & 0.9846 & 0.9349 & 0.9597 & 0.9824 & 0.9334 & 0.9579 & 0.9569 & 0.8985 & \textbf{0.9277} & 0.9484 \\
\hline
Text deformer & 0.8641 & 0.8363 &0.8502& 0.8630 & 0.8425 &0.8528& 0.8534 & 0.8173 & 0.8353 & 0.8461  \\
X-mesh & 0.8626 & 0.8602 &  0.8614& 0.8817 & 0.8950 &0.8884& 0.8635 & 0.8448 &  0.8541& 0.8680 \\
Deformation transfer & 0.9768 & 0.9376& 0.9572 & 0.9598 & 0.9576 & 0.9587& 0.9254 & 0.9288 & 0.9271 & 0.9477\\
\hline
\end{tabular}
\label{Tab:qualitative}
\end{table*}

\paragraph{Comparison with Baselines on Mesh Stylization}
\label{sec:qualitative}

We compared our stylized face generation results with those produced by baselines that can deform a mesh into different styles: Deformation Transfer~\cite{sumner2004deformation}, TextDeformer~\cite{gao2023textdeformer}, and X-mesh~\cite{ma2023x}. In case of Deformation Transfer~\cite{sumner2004deformation}, we identified the correspondences \revised{between the identity exemplar mesh and the deformation target using 68 facial landmarks along with 9 additional points on the forehead in order to apply learned deformation.} TextDeformer and X-mesh employ a text-guided deformation that operates on an input mesh. \fixed{For both methods, the input face is deformed based on the same descriptive text to generate stylized outputs.}
\fixed{All these baselines were evaluated on three different mesh topologies: FLAME and FLAME with Loop subdivision and mesh simplification.}

\begin{figure}[h]
\centering
\includegraphics[width=1\linewidth]{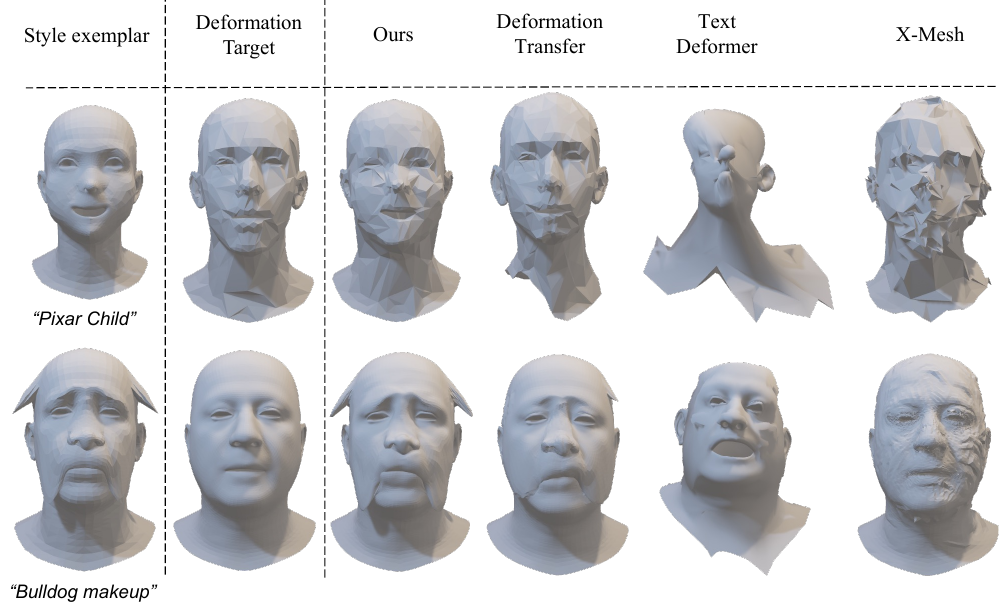}
\caption{Qualitative comparison on stylization. Our method adheres to style and identity without artifacts unlike Deformation Transfer. Text-based approaches fail to match styles from text.}
\vspace{-5pt}
\label{comparison}
\end{figure}

We present qualitative comparison results on stylized 3D face generation in Figure~\ref{comparison}. Deformation Transfer is capable of generating faces that follow the style; however, it exhibits artifacts, including anomalies and concavities on the surface due to the calculation of displacement and its direct transfer. In contrast, text-based methods fall short in following styles. Our method, on the other hand, can generate stylized 3D faces that adhere to both the \fixed{input} identity and style example without any artifacts. 

In quantitative evaluation, shown in Table~\ref{Tab:qualitative}, our method obtained the highest average scores computed from CLIP-SP and CLIP-IP. This clearly demonstrates that our approach achieves better performance in mesh stylization while preserving input identity compared to deformation baselines. More details and examples are provided in the supplementary material.

\vspace{-3mm} 
\paragraph{Surface-Intensive Mesh Sampling} \label{exp:SIMS}
To test the proposed sampling method SIMS, we conducted a study on the reconstruction task \fixed{comparing three different sampling variants} for training $D_S$. SIMS is surface-intensive sampling, using approximately four times more points sampled from the mesh surface \fixed{compared to mesh vertices.} Hybrid sampling, originally proposed in DD3C~\cite{jung2022deep}, sampled points both vertices and faces in similar ratio ($\sim$1.1 times). Vertex-only used just the vertices of the mesh.

For the experiment, we calculated the reconstruction loss on four different experiment settings: \fixed{(1) original FLAME, (2) a simplified FLAME mesh with 1/4 the vertex count, (3) FLAME with one loop subdivision, and (4) FLAME with two loop subdivisions.} The reconstruction loss was compared to the ground truth position using mean squared error. As shown in Table~\ref{Tab:SIMS}, SIMS achieved the lowest error by a large margin on all experiments. Hybrid sampling generally performed better than vertex-only sampling.

\begin{table}[t]
\centering
\renewcommand{\arraystretch}{0.9} 
\caption{Reconstruction comparison with mesh sampling methods}
\vspace{-10pt}
\small
\setlength{\tabcolsep}{3pt} 
\begin{tabular}{|c|ccccc|}
\noalign{\smallskip}\noalign{\smallskip}\hline
& \multicolumn{5}{c|}{Reconstruction loss↓ (all in e-5)} \\ \hline
Methods & original & simplified & loop-1 & loop-2 & average \\ \hline
SIMS(Ours) & \textbf{1.790} & \textbf{1.635} & \textbf{1.612} & \textbf{1.591} & \textbf{1.657}  \\
Hybrid & 2.300 & 2.019 & 2.138 & 2.118 & 2.144 \\
Vertex & 1.955 & 5.866 & 3.258 & 2.959 & 3.509  \\
\hline
\end{tabular}
\label{Tab:SIMS}
\end{table}

\begin{figure}[h]
\centering
\includegraphics[width=1\linewidth]{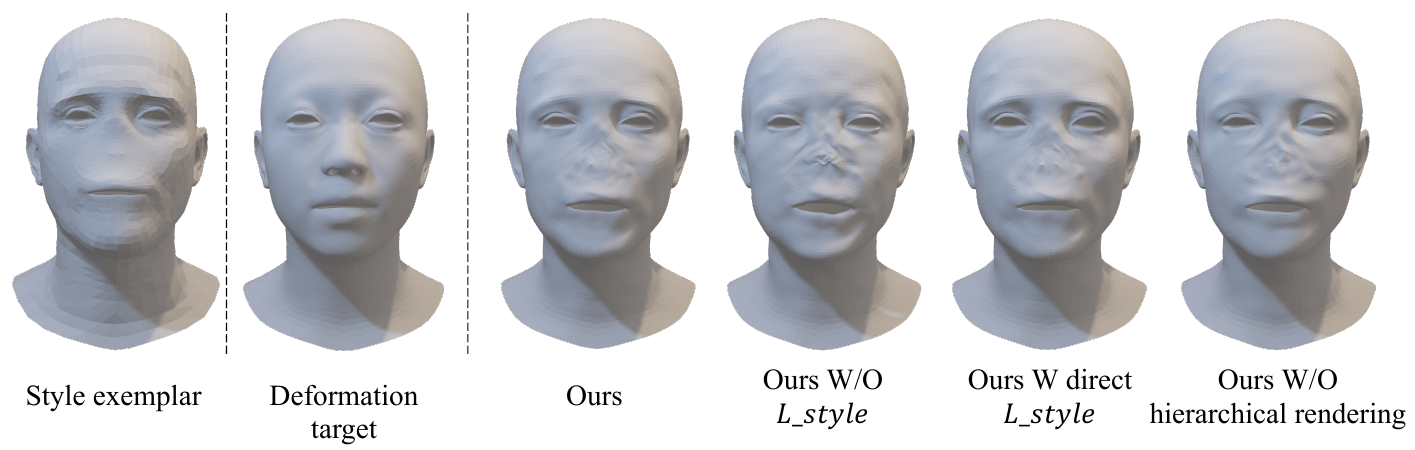}
\caption{Visualization on ablation study.}
\vspace{-5pt}
\label{fig:ablation}
\end{figure}

\paragraph{Ablation Study}
To perform an ablation study, we trained our model $D_T$ using different settings: 
1) ``Ours w/o $L_{style}$," where the style loss was removed; 2) ``Ours w direct $L_{style}$," where the style loss was directly compared between \fixed{$D_T([{z_s}^{samp};{z_e}^{samp}])$} and $M_T$; 3) ``Ours w/o hierarchical rendering," where a single canonical view was used to calculate the $L_{in}$ and $L_{across}$. The results are displayed in Figure~\ref{fig:ablation} and Table~\ref{Tab:qualitative}. The results produced by "Ours" and "Ours w direct $L_{style}$" do not show any artifacts unlike the results produced by w/o $L_{style}$ and w/o hierarchical rendering, which exhibit surface artifacts. However, ``Ours w. direct $L_{style}$" forces \fixed{$D_T([{z_s}^{samp};{z_e}^{samp}])$} and $M_T$, which have different expressions, to have the same normals; thus, it discards the expression. As a result, while ``Ours w. direct $L_{style}$" does stylize effectively, it cannot animate. Conversely, ``Ours" not only stylizes well but is also animatable. Additional ablation study results regarding animation are presented in the supplementary material.

\vspace{-2mm}

\vspace{-4mm}  
\paragraph{User Study}
We conducted a user study with 34 participants to evaluate different stylized face generation methods on human perception. A total of eight different styles were used. Each participant was presented with 20 questions and asked to choose the result that best followed the style while preserving the identity. As shown in Table~\ref{tab:perceptual}, our method received higher scores compared to the baseline methods. Ours outperformed the text-based stylization methods by a large margin and scored higher than the method requiring correspondence specification.

\begin{table}[t]
\centering
\renewcommand{\arraystretch}{0.9} 
\caption{Results from the user perceptual study given style exemplar and the deformation target. The percentage represents the selected frequency.}
\vspace{-5pt}
\footnotesize
\renewcommand{\arraystretch}{0.9} 
\label{tab:perceptual}
\begin{tabular}{|l|c|}\hline 
\multicolumn{1}{|c|}{\textbf{Method}} & User Score     \\ \hline
Ours                            & \textbf{ 60.65\%} \\ 
Deformation transfer                   & 38.17\%\\
Text deformer                          &  0.44\%\\ 
X-mesh                                 &  0.74\%\\ 
\hline 
\end{tabular}
\vspace{-0.5cm}
\end{table}
\section{Applications}
\label{sec:application}

\subsection{Style Interpolation}
\label{app:style_interpolation}
\begin{figure}[h]
\centering
\includegraphics[width=1\linewidth]{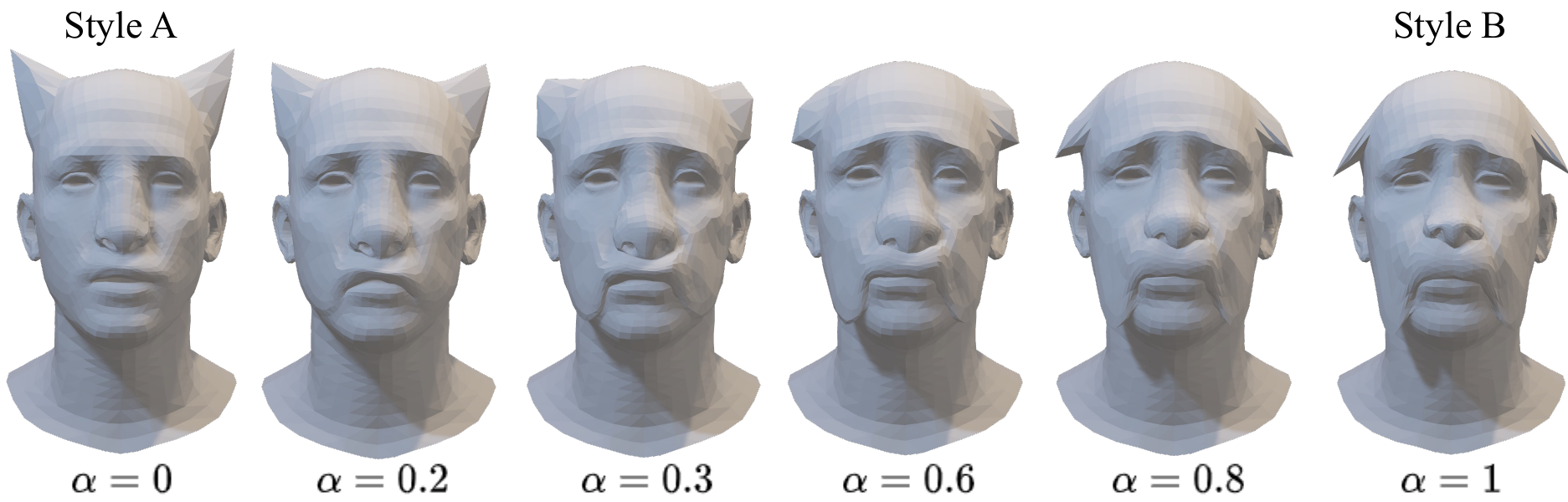}
\vspace{-2mm}
\caption{Linear interpolation of Style A and Style B}
\label{app1}
\vspace{-4mm}
\end{figure}

LeGO enables linear style interpolation as illustrated in Figure~\ref{app1}, through weight blending. Diverse styles are seamlessly blended to create new stylized meshes by linearly interpolating weights from $D_{T_1}$ and $D_{T_2}$ using following equation:
\vspace{-2mm} 
\begin{equation}
\label{eq:inter}
W_{\text{new}} = \alpha W_{D_{T_1}} + (1-\alpha) W_{D_{T_2}} 
\end{equation} 

Here, $W_{D_{T_1}}$ and $W_{D_{T_2}}$ denote the \revised{network} weights for $D_{T_1}$ and $D_{T_2}$, respectively, while $\alpha$ represents the blending weight controlling the interpolation. 

\subsection{Image-based 3D Stylized Avatar Generation}
\label{app:reconstylized}
\begin{figure}[h]
\centering
\vspace{-1mm}
\includegraphics[width=0.9\linewidth]{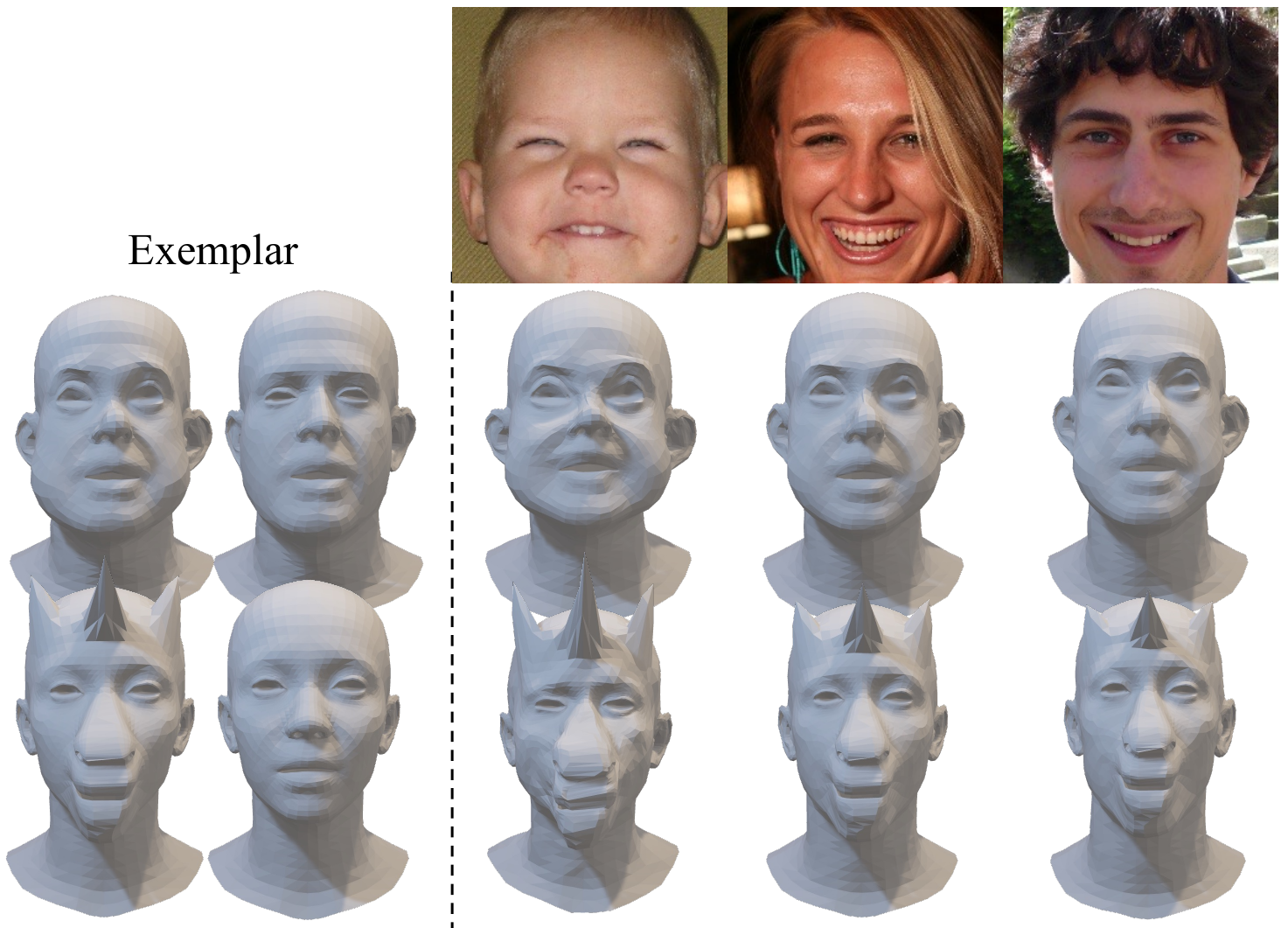}
\vspace{-1mm}
\caption{Visualization of generating stylized 3D faces from 2D portraits~\cite{karras2019style}.}
\vspace{-3mm}
\label{app2}
\end{figure}

A stylized 3D face can be generated from a single portrait by first using methods that reconstruct 3D faces from 2D portraits. Among these methods, we employed MICA~\cite{zielonka2022towards} to reconstruct a 3D face from an image. This reconstructed shape was then fed into LeGO to create stylized faces. The results are visualized in Figure~\ref{app2}.
\vspace{-3mm} 
\section{Limitation and Conclusion}

\fixed{We presented a novel approach for generating stylized 3D face meshes, considering three key elements. First, we proposed a surface deformation network that can generate a face in the desired topology using SIMS. Second, by the domain adaptation with hierarchical rendering, we achieved superior stylization capability. Lastly, Using 3DMM prior, we can generate a stylized face equipped with the animation capability.}

In addition, we proposed MAGE for practical usage, which can take diverse mesh topologies as input, and a novel style loss that adheres to the style effectively while preserving animation ability. Comprehensive experimental results demonstrate that our method is capable of generating a stylized mesh with consistent topology given deformation target meshes exhibiting significant topological variation.

While our method shows promising results and enables significant advancements in practical avatar creation, it also has challenges to address. At inference, achieving a stylized output with the same topology as the input requires a two-stage process involving template replacement with the mean face mesh. Addressing these efficiency and practicality challenges is crucial for further enhancing our approach.
\clearpage
{
    \small
    \bibliographystyle{ieeenat_fullname}

}


\end{document}